\documentclass[lettersize,journal,compsoc]{IEEEtran}
\usepackage{amsmath,amsfonts}
\usepackage{algorithmic}
\usepackage{algorithm}
\usepackage{array}
\usepackage{textcomp}
\usepackage{stfloats}
\usepackage{url}
\usepackage{verbatim}
\usepackage{cite}
\usepackage{booktabs}
\usepackage[table,xcdraw]{xcolor}

\usepackage{balance}
\usepackage{multirow}
\usepackage{graphicx}
\usepackage{hyperref}

\usepackage{subcaption}

\begin{document}
\newcommand\textTitle{The Effects of Embodiment and Personality Expression on Learning in LLM-based Educational Agents}

\title{\textTitle}

\author{Sinan Sonlu, Bennie Bendiksen, Funda Durupinar, U\u{g}ur G\"{u}d\"{u}kbay,~\IEEEmembership{Senior~Member,~IEEE} %
\IEEEcompsocitemizethanks{
\IEEEcompsocthanksitem Manuscript received June 19, 2024; revised August 16, 2024.
\IEEEcompsocthanksitem S. Sonlu and U. G\"{u}d\"{u}kbay are with the Department of Computer Engineering, Bilkent University, Ankara, Turkey. e-mails: sinan.sonlu@bilkent.edu.tr, gudukbay@cs.bilkent.edu.tr.
\IEEEcompsocthanksitem B. Bendiksen and F. Durupinar are with the Department of Computer Science, The University of Massachusetts Boston, USA. e-mails: B.Bendiksen001@umb.edu, Funda.DurupinarBabur@umb.edu.
}}     % <-this % stops a space


\markboth{\textTitle}
{Sonlu \MakeLowercase{\textit{et~al.}}: \textTitle}


\IEEEpubid{0000--0000/00\$00.00~\copyright~2021 IEEE}

\maketitle

\begin{abstract}
%todo
This work investigates how personality expression and embodiment affect personality perception and learning in educational conversational agents. We extend an existing personality-driven conversational agent framework by integrating LLM-based conversation support tailored to an educational application. We describe a user study built on this system to evaluate two distinct personality styles: high extroversion and agreeableness and low extroversion and agreeableness. For each personality style, we assess three models: (1) a dialogue-only model that conveys personality through dialogue, (2) an animated human model that expresses personality solely through dialogue, and (3) an animated human model that expresses personality through both dialogue and body and facial animations. The results indicate that all models are positively perceived regarding both personality and learning outcomes. Models with high personality traits are perceived as more engaging than those with low personality traits. We provide a comprehensive quantitative and qualitative analysis of perceived personality traits, learning parameters, and user experiences based on participant ratings of the model types and personality styles, as well as users' responses to open-ended questions. 

\end{abstract}


\section{Introduction}
\IEEEPARstart{V}{irtual} humans have tremendous potential to support educational activities by providing on-demand, personalized learning experiences. Their function is more than just relaying information; they can socially connect with users, establish rapport, and motivate them~\cite{swartout2013virtual}. Their potential has significantly increased with the advancement of Large Language Models~(LLMs), which can effectively assume various roles and personalities and provide information on any topic. Virtual humans with LLM-driven dialogue capabilities offer customized experiences for learners with diverse preferences and needs.   

A substantial body of literature examines how teacher personalities impact effective learning and student preferences~\cite{murray1990teacher, furnham2005individual, chamorro2008birds, kim2016students}.   Overall, all the Five-Factor Model (FFM)~\cite{costa1992five} personality traits with positive connotations--openness, agreeableness, conscientiousness, extroversion, and emotional stability (negative neuroticism)--play important roles in educational contexts~\cite{tan2018students}. However, factors such as individual differences among students~\cite{furnham2005individual} and the type of the course~\cite{murray1990teacher} determine which traits are effective in specific scenarios.

Research has shown that embodied characters are perceived as more trustworthy, engaging, and socially present than their disembodied counterparts~\cite{Kim2018}. Embodied pedagogical agents have been reported to increase motivation and enjoyment in learning; yet, their effects on knowledge acquisition have been mixed~\cite{schroeder2013effective, petersen2021pedagogical} as they also increase cognitive load and cause distraction.  

Building on these insights, we investigate how virtual agents' embodiment and personality expression affect learning outcomes in an educational application.   This work extends an existing personality-driven conversational agent framework~\cite{sonlu2021conversational} with LLM-based conversation support tailored for an educational scenario.  In our application, users interact with a conversational agent in real time by typing their questions, to which the agent responds in speech vocalized by the operating system's text-to-speech functionality.

As the system's personality parameters, we chose the combination of extroversion and agreeableness since previous work showed that these two personality factors were more effectively conveyed through body gestures and facial expressions than the other three~\cite{sonlu2021conversational}. We refer to them as an agent's ``personality style'' with high-trait and low-trait variants, corresponding to a high extroversion-agreeableness combination and a low extroversion-agreeableness combination. The former is manifested as a friendly, vivid, and energetic agent designed to engage users with enthusiasm and warmth. In contrast, the latter is characterized by a more reserved and less approachable demeanor, offering interactions that may be perceived as less engaging and more formal. Our system expresses personality through dialogue text and/or body and face animation. 

We created three models to assess embodiment: a dialogue-only model and two models with 3D humanoid bodies. All the models displayed conversation text concurrently with audio feedback. We evaluated the efficacy of different modalities and personality styles on learning through an independent-subjects user study. The study randomly presented each participant with a high or low personality variant of each model.  The dialogue-only model and one of the embodied models expressed personality only through text, and the other embodied model expressed personality through face and body movements and gaze.  During the study, we collected ratings about the system for learning, quality, and engagement as well as the perceived personalities of the agents. Additionally, we obtained user feedback through open-ended questions.  
 
Although the system parameters were selected to express the personality dimensions of extroversion and agreeableness, we considered potential variations in participants' perceptions of the agents' personalities. For instance, a high-trait agent might also be perceived as emotionally stable with low neuroticism, or a dialogue-only agent might be viewed as conscientious, even though these traits were not intentionally highlighted. Therefore, we collected users' perceptions of the agents' personalities across all five dimensions of the FFM for each of the three models.

We aim to answer the following research questions:
\begin{itemize}

  \item [\textbf{RQ1.}] Is there an effect of personality style on perceived personality?  
  
  \item [\textbf{RQ2.}] Is there an effect of model type on perceived personality? 

  \item [\textbf{RQ3.}] Is there an effect of model type on learning outcomes?

  \item [\textbf{RQ4.}] Is there a correlation between learning outcomes and personality perception?

\end{itemize}

Based on the findings of the previous studies, we have the following hypotheses: 
\begin{itemize}
   \item [\textbf{H1.}] Learning outcomes will be rated higher for the embodied agents than the dialogue-only agent, reflected as higher ratings in H1a. learning; H1b. quality; H1c. engagement. Since the literature indicates a more positive approach towards embodied agents than disembodied ones, we expect a similar tendency in our educational application. In other words, the embodied agents will be more engaging and effective in learning~\cite{petersen2021pedagogical}.
  
  \item [\textbf{H2.}] Agents expressing high extroversion and agreeableness will be rated higher in terms of learning outcomes than the agents expressing low extroversion and agreeableness, reflected as higher ratings in H1a. learning; H1b. quality; H1c. engagement. We formulate this hypothesis based on students' reported preferences for teacher personalities and documented preferences of users for highly agreeable chatbots~\cite{volkel2021examining, mehra2021chatbot}.
  % We expect high-trait agents to be more engaging than low-trait agents, leading to higher-rated learning outcomes.
\end{itemize}

In addition to investigating these questions through quantitative analysis, we identify common themes and individual differences across participants by an in-depth qualitative analysis of their responses to open-ended questions. Furthermore, we provide our system as an open-source virtual tutoring application with conversational virtual agents that exhibit desired personality traits via motion and language. Our data and code are available in our public repository~\footnote{Repository link will be available in the final version.}

\section{Related Work}
Involving multiple computing fields, this work on educational agents combines personality expression, LLM-based dialogue generation, and conversational agent systems. We summarize the related work based on these categories, including similar studies on pedagogical agents.

\subsection{Pedagogical Agents}
The earliest usage of educational computer software includes military applications such as flight simulations~\cite{aebersold2016history}. Increasing widespread use of computers opened the way for many interactive multimedia applications focusing on education~\cite{troutner1991historical}. One special form of computer-assisted learning includes life-like pedagogical agents that help with learning and motivation in multimedia environments~\cite{heidig2011pedagogical}. Such agents can assume the role of instructors, coaches, tutors, or learning companions~\cite{gulz2011building} and converse with the user using natural language with text or speech input~\cite{weber2021pedagogical}. Pedagogical agents can simulate instructional roles such as expert, motivator, and mentor with high accuracy~\cite{baylor2005simulating}.  Human-like agents have been shown to influence learner achievement, attitude, and retention of learning~\cite{yilmaz2012educational} and deliver a more interesting overall learning experience than learning without an agent~\cite{lin2020using}.

Research has explored the use of 2D and 3D agents in educational settings, revealing varying effects on learning outcomes. 2D agents can outperform their more visually complex 3D counterparts in certain scenarios~\cite{castro2021effectiveness}, as they reduce extraneous cognitive load and enhance coherence in multimedia learning~\cite{mayer201412}. High levels of anthropomorphism in agents can detract from social co-presence~\cite{nowak2003effect} and consequently limit their pedagogical benefits~\cite{zhang2022benefits}. 3D agents, in contrast, offer a greater variety of expressions and gestures, which can boost learning and engagement~\cite{davis2018impact}. Although high behavioral realism enhances social presence in virtual reality environments, where 3D agents are particularly effective, it can negatively impact factual learning outcomes~\cite{petersen2021pedagogical}.  However, the presence of a real instructor is associated with increased learning and satisfaction, leading to better information recall~\cite{wang2017instructor}. The mixed effects of high behavioral realism may be explained by the uncanny valley effect~\cite{mori2012uncanny}, where realistic yet not quite human-like agents induce discomfort and unease. In robotics, incorporating personality into humanoid robots has been shown to reduce these uncanny feelings and enhance the overall user experience~\cite{paetzel2021influence}, which can also apply to virtual humans.

% ~\cite{tao2022exploring}

Pedagogical agents can be embedded into different applications such as text-based mobile conversational systems~\cite{kloos2018design}, collaborative serious games~\cite{terzidou2014impact}, multiple-agent intelligent tutoring systems~\cite{lippert2020multiple}, collaborative augmented reality environments~\cite{zielke2024exploring}, and artificial intelligence-enabled remote learning\cite{atif2021artificial}.  In robotic educational agents, verbal cues are more effective than nonverbal cues in improving engagement~\cite{brown2013applying}. Expressing various emotions through facial expressions and body motion can appeal to different learner types in animated pedagogical agents~\cite{adamo2021multimodal}.  Pedagogical agents have been utilized in teaching a wide range of subjects, including STEM~\cite{terracina2016teaching}, foreign language~\cite{carlotto2016effects}, history~\cite{poitras2014developing}, as well as work training~\cite{khokhar2022modifying}. Learning, engagement, human likeness, credibility, and personality can be used as measures of the psychometric structure of pedagogical agents~\cite{ryu2005psychometric}.

% Not all pedagogical agents assume the role of the teacher; users can teach the agent through organizing the reasoning structures of the domain in an educational setting that can improve the user's learning outcomes~\cite{biswas2005learning}, or cooperate with a remote-controlled agent to identify countries on the world map in a pedagogical reference resolution game~\cite{paetzel2020rdg}.

\subsection{Personality Expression in Agents}
Multi-modal communication elements are essential for expressing desired personality traits in digital characters. Nonverbal behavior elements are commonly used in affective virtual agents to convey these traits~\cite{perlin1995personality, andre2000exploiting, saberi2016computational, saberi2021expressing}. Studies leverage high-level meanings of motion to express the target personality type~\cite{allbeck2002toward}. For instance, PERFORM establishes a link between Laban Movement Analysis~(LMA) parameters and the perceived personality of virtual human characters~\cite{durupinar2016perform}. In addition to body motion,  personality-specific voice, dialogue, and facial expressions help distinguish personality traits in expressive conversational agents~\cite{sonlu2021conversational}. Recent research indicates that appearance and movement significantly influence the expression of certain traits such as agreeableness and neuroticism~\cite{yurtouglu2024personality}.

The character's rendering style also affects personality perception~\cite{zell2019perception}. For example, cartoon-like rendered characters are perceived as more agreeable, whereas characters with unappealing, ill-looking rendering styles are found quarrelsome and less sympathetic~\cite{zibrek2014does}. Additionally, animation realism~\cite{thomas2022investigating}, face model~\cite{branham2001creating}, body shape~\cite{swami2008five, hu2018first}, clothing, environment, and facial expression~\cite{legde2019evaluating}, as well as skin texture and viewing angle~\cite{jones2012signals} of virtual characters are all influential on perceived personality. In multi-agent scenarios, the interaction between agents and proximity is a successful indicator of personality~\cite{durupinar2009ocean, kapadia2013authoring, durupinar2015psychological}, as well as emotional group dynamics~\cite{durupinar2015psychological}. Action choices in procedural story generation can also express different personality types~\cite{affectivestory2007, bahamon2017empirical}. An agent's perceived personality highly influences how users interact with the system; for example, users are more willing to trust and listen to serious-looking, assertive agents~\cite{zhou2019trusting}.

Human hands are highly expressive in communication~\cite{adkins2023important}; specific hand movements can convey different personalities~\cite{wang2016assessing}. Gesture performance in combination with language highly influences perceived personality~\cite{neff2010evaluating}. Linguistic elements such as the ratio of phrases, words of emotion and cognition, and exclamations correlate with apparent personality traits~\cite{mairesse2009can, lee2007relations}. Automatic personality assessment is possible using text input from social media messages~\cite{personality2011golbeck, park2015automatic}, using speech~\cite{polzehl2010automatically,speech2010moller}, facial expressions~\cite{biel2012facetube}, gait features~\cite{gaitpersonality2018} and body motion features~\cite{sonlu2024towards}. Overall, different communication elements contribute to the perceived personality of digital characters where multi-modal approaches often yield better results~\cite{vinciarelli2014survey, skowron2016fusing, kampman2018investigating}.

LLMs such as GPT exhibit consistent personality cues~\cite{huang2023revisiting} and can be customized for different personalities through prompting~\cite{gu2023effectiveness}, as we do in this work. Analysis suggests that word choices and the length of the generated text are influential on this~\cite{safdari2023personality}. Data-driven personality estimation systems can predict different personality types when the generated text uses certain prompts~\cite{mehta2020bottom, karra2022estimating}, supporting the success of LLMs in capturing personality cues from language. 

\subsection{LLM-Based Agents}

LLMs span many applications, including natural language-based human-computer interaction~\cite{huang2023revisiting}. The rise of LLMs has opened up new avenues for creating and populating digital worlds.   For instance, language models can be used to generate and animate 3D avatars~\cite{hong2022avatarclip}, schedule motions~\cite{qing2023story}, control facial expressions and body motion styles~\cite{normoyle2024using}, drive non-player character behavior in games~\cite{kumaran2023scenecraft, normoyle2024using}, and automate and refine digital storytelling~\cite{sohn2024words}. 3D scenes can be injected into LLMs for captioning, 3D question answering, and navigation tasks~\cite{hong20243d}; models can describe or compare objects in 3D scenes~\cite{wang2023chat}. 

LLMs have also started playing critical roles in innovative educational technologies~\cite{kasneci2023chatgpt}. For instance, LLMs are used with vocalized agents in Augmented Reality (AR) environments to aid students in foreign language learning~\cite{topsakal2022framework}. Similarly, in healthcare, LLMs enhance patient experiences during consultation, diagnosis, and management~\cite{yang2023large}. Although dialog systems utilizing LLMs can generate highly sophisticated responses, they lack access to dynamic real-time data, such as the current date~\cite{villa2023conversational}. Consequently, LLM-based agent systems often focus on isolated tasks, such as answering questions based on pre-existing knowledge. To enable more context-aware conversations, additional mechanisms to enhance LLM memory can be implemented~\cite{liu2024llm}. In this work, we use an LLM to generate educational responses for specific computing subjects in an educational setting.

\subsection{Conversational Agents}
Conversational agents aim to interpret and respond to user statements in ordinary natural language through integrating computational linguistics techniques~\cite{lester2004conversational}. Understanding and exhibiting emotions and personality are essential for successful natural language conversations~\cite{ball2000emotion}. For example, the same query may require different interpretations based on the user's current mood, and similarly, the same response can be perceived differently based on the agent's current body language and facial expression. Conversational systems that give relevant answers to user questions are perceived as more human-like and engaging~\cite{schuetzler2018investigation}. The realism of animation and behavior is critical for agents with a visual representation \cite{thiebaux2008smartbody}. Thus, studies synthesize gesture animation to accompany speech~\cite{kopp2004synthesizing, nyatsanga2023comprehensive}. Recent deep learning-based co-speech animation generation systems can produce highly realistic results~\cite{qian2021speech, bhattacharya2021,  ao2022rhythmic, bhattacharya2024speech2unifiedexpressions}, and the generated animations can be authored to express the desired pose and style at specific frames~\cite{yoon2021sgtoolkit}.  Facial expressions are also crucial to create realistic experiences. For instance, agents that mimic user facial expressions deliver believable and empathetic conversations~\cite {aneja2021understanding}.  Conversational agents can take as input multiple stimuli, including user's gaze~\cite{ishii2013gaze}, speech and facial expression~\cite{aneja2021understanding}, structured or natural language text~\cite{tudor2020conversational}. In this study, we input natural language text from participants to keep the system requirements minimal and leave other modalities for future work.

\section{Method}
This section describes a user study to test the impact of different modalities and personalities on learning and personality perception factors. For this, we designed an application employing a conversational agent that teaches a complex topic of the user's choice through turn-based dialogue~\footnote{Bilkent University Ethical Committee for Human Research approved the study with the decision number 2023\_11\_05\_01.}.

\subsection{System}

\begin{figure*}[htbp]
    \includegraphics[width=\textwidth]{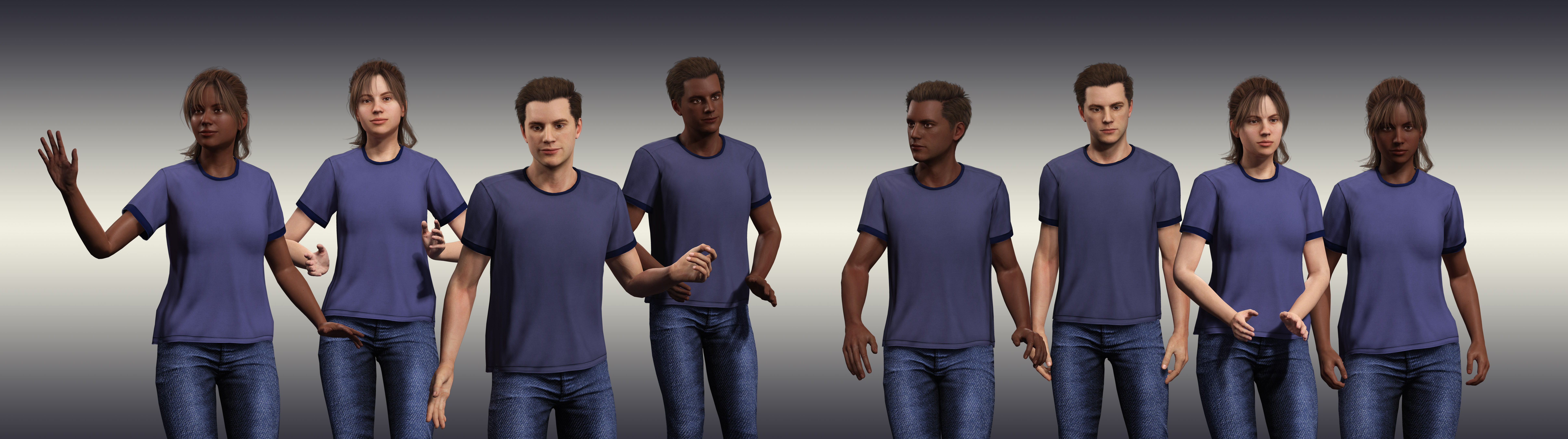}
    \caption{Different 3D agent models used in the study expressing high (left group) and low (right group) traits.}
    \label{fig:agentmodels}
\end{figure*}

To run our study, we updated the personality-driven conversational agent platform by Sonlu~et~al.~\cite{sonlu2021conversational}, an open-source, multi-modal system for animating 3D conversational virtual agents. The platform runs on Unity~\cite{Unity} and controls multiple modalities, such as dialogue, facial expressions, and body movements, based on an input personality. Body movement control involves modifying a given animation clip via joint rotation and animation speed adjustments, noise addition, and inverse kinematics-based gesture changes following the LMA mappings defined in PERFORM~\cite{durupinar2016perform}. Facial animation involves mouth movements during speech, frequent blinks associated with neuroticism, and updating blend shapes to express emotions associated with specific personality factors. Facial expressions are designed according to Facial Animation Control System (FACS)~\cite{Ekman2002}, and mouth movements are handled by Oculus LipSync~\cite{oculuslipsynch} with a customized mapping to the facial blend shapes of the 3D model. The input personality determines the agent's default facial expression. For example, an agreeable agent tends to smile by default with each turn of its dialogue, which diminishes with time. We designed 3D human models for the current study using Reallusion Character Creator~\cite{Reallusion4}. To introduce a measure of diversity, we created four characters: two female and two male, each with light and dark skin tones, as depicted in Figure~\ref{fig:agentmodels}.

Our updated system differs from the existing personality-driven platform in terms of how it handles dialogue. The previous work used IBM Watson Assistant~\cite{WatsonAPI} to extract the intent from user queries mapped into domain-specific handcrafted dialogue lines. This work replaces the dialogue logic with an LLM-based text generation model, GPT-3.5 Turbo~\cite{openai:gpt3.5}, facilitated by OpenAI's Chat Completions Application Programming Interface (API), eliminating the need for manual dialogue crafting. When users type their prompts (e.g., ask a question), the system returns an answer coherent with the input personality description. We limit the token number to 750 for the output text to keep the conversation concise. Unlike the previous platform, which used Watson Text-to-Speech API for speech generation, our current system employs Microsoft\textregistered \ text-to-speech functionality, producing an almost immediate response to vocalize the agent's answer. This local solution also lets us determine the currently spoken word we use to display partial subtitles. Since the generated responses could be fairly long, we followed a dynamic approach where five words centering the currently spoken word were displayed on top of the agent in models with visual representation. The subtitles were on by default, but the users could disable them in models with 3D agents if they found them distracting. 

We used a temperature of 0.9 to promote diverse outputs from GPT while maintaining the information's reliability. Temperatures greater than 1 introduce creativity; however, they lead to hallucinations, conflicting with the aim of the educational system. Chat Completions API takes as input a ``messages'' parameter consisting of message objects, where each object has a role of ``system'', ``user'', or ``assistant'' and content. For the role of ``system'', we give the following messages as input for different agent personalities and teaching topics:
\begin{itemize}
  \item\noindent {\{``role'': ``system'', ``content'': Act as an extroverted teacher teaching about $<$\textit{topic}$>$, give friendly and polite answers.\}}
   \item\noindent {\{``role'': ``system'', ``content'': Act as an introverted teacher teaching about $<$\textit{topic}$>$, give short and unfriendly answers.\}} 
\end{itemize}

To produce a response that the agent speaks, the system sends the role prompt together with the last five dialogue messages, alternating between the user and assistant: {\{``role'': ``user~\slash~assistant'', ``content'': $<$ \textit{message}~$>$\}} We limited the number of messages to five to reduce costs, eliminate context drift, and prevent manipulation. This restriction helps prevent altering the language model's perception through extended interactions~\cite{dynel2023lessons}.

\subsection{Stimuli}
We designed a $3 \times 2$ independent subjects study for a comparative analysis of three models—D, A, and E—each tested with high and low values of agreeableness-extroversion combination. Model D is the \underline{d}ialogue-only setting, where the system's answers were shown on screen sentence-by-sentence concurrently with audio playback. Model A included an \underline{a}nimated 3D model of the agent, randomly chosen among four alternatives (two male, two female, each with dark and light skin tones). Model A involved the virtual human animated without any personality-based alterations. In models D and A, personality was conveyed only through synthesized dialogue. Model E was similar to Model A but additionally incorporated the \underline{e}xpression of personality through face and body movements.  

In model E, motions that display a combination of high extroversion and agreeableness involve the LMA parameters of Indirect Space, Light Weight, and Free Flow~\cite{durupinar2016perform}. These correspond to multi-focal spatial attention, delicate, lifted-up movements, and uncontrolled and fluid motion. Because high extroversion and agreeableness are associated with opposite Time Efforts (Sudden, urgent vs. Sustained, lingering), we left the Time component of the animations unaltered. The animations expressing low extroversion and agreeableness involve Direct Space, Strong Weight, Bound Flow, and neutral Time, corresponding to single-focused, heavy, and controlled movements. The default facial expression of a highly extroverted and agreeable agent is relaxed and happy, with occasional smiles and direct eye contact. In contrast, an agent characterized by lower levels of these traits displays a more tense expression by default. Low-trait agents also avoid eye contact with the viewer.

\begin{figure*}[htbp]
   \centering
   \begin{subfigure}{0.48\textwidth}
     \centering
     \includegraphics[width=\textwidth]{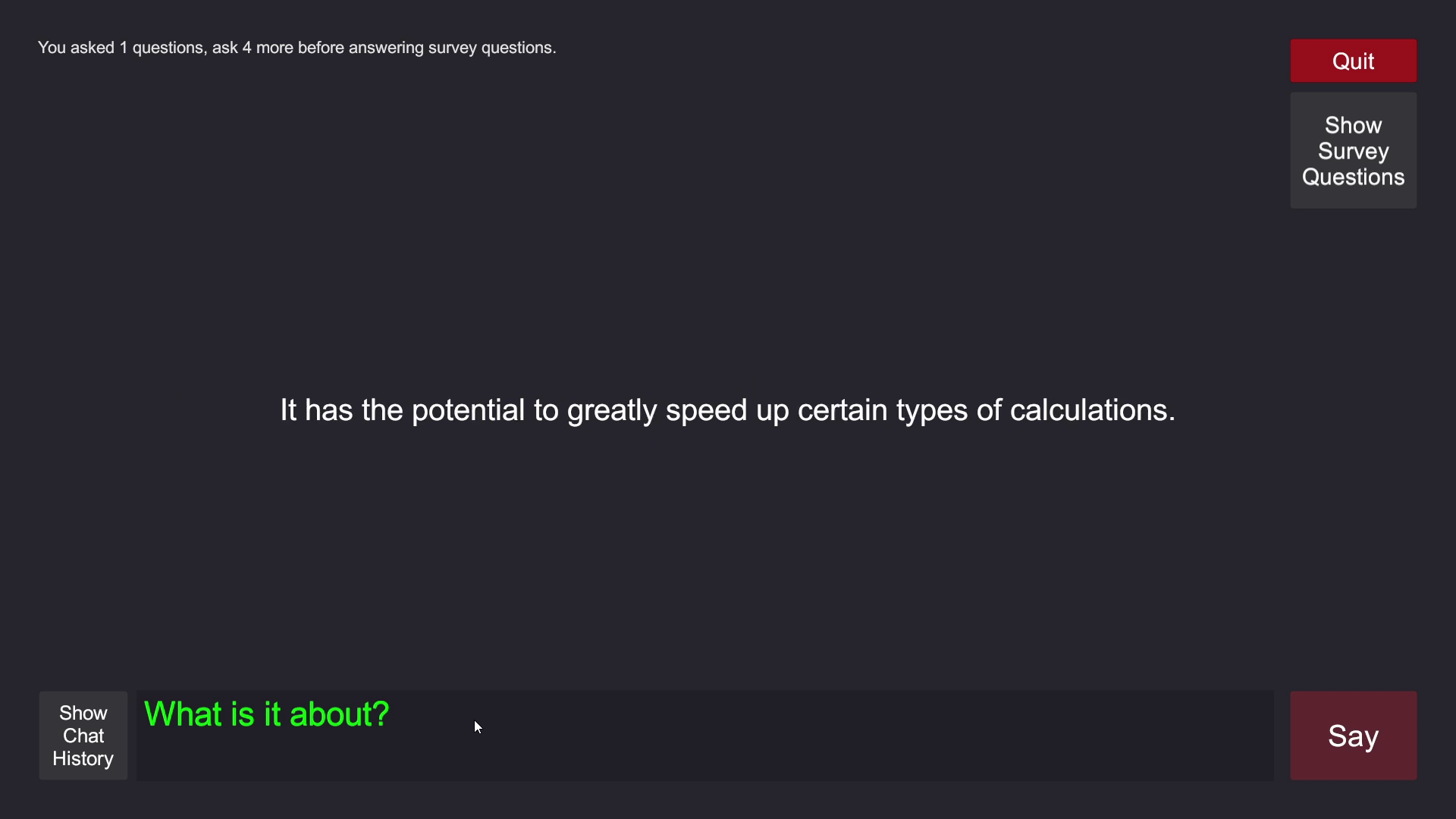}
     \caption{Model D}
     \label{fig:model-t}
   \end{subfigure}
   \begin{subfigure}{0.48\textwidth}
     \centering
     \includegraphics[width=\textwidth]{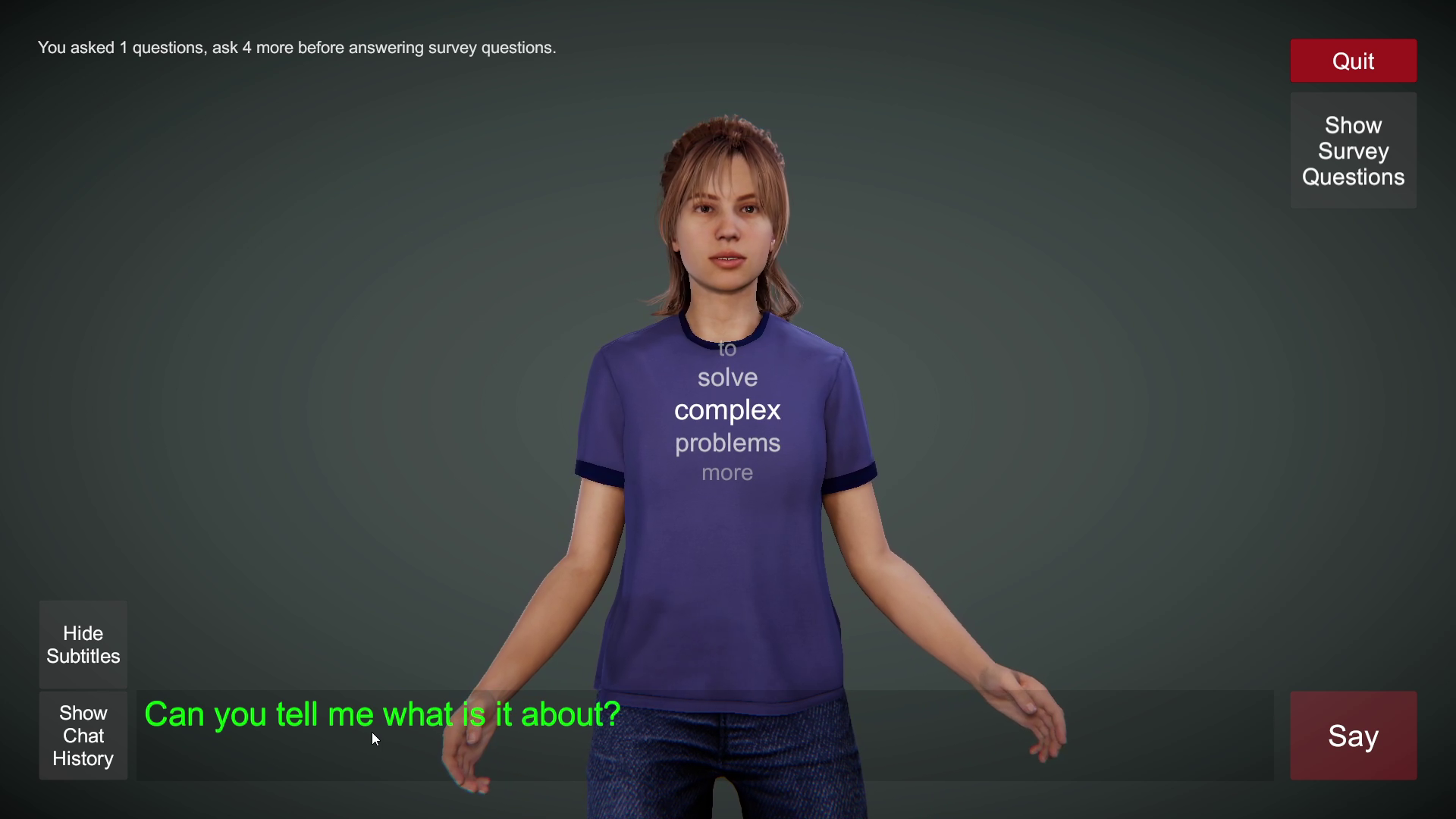}
     \caption{Model A}
     \label{fig:model-a}
   \end{subfigure}
   %\hfill
   \begin{subfigure}{0.48\textwidth}
     \centering
     \includegraphics[width=\textwidth]{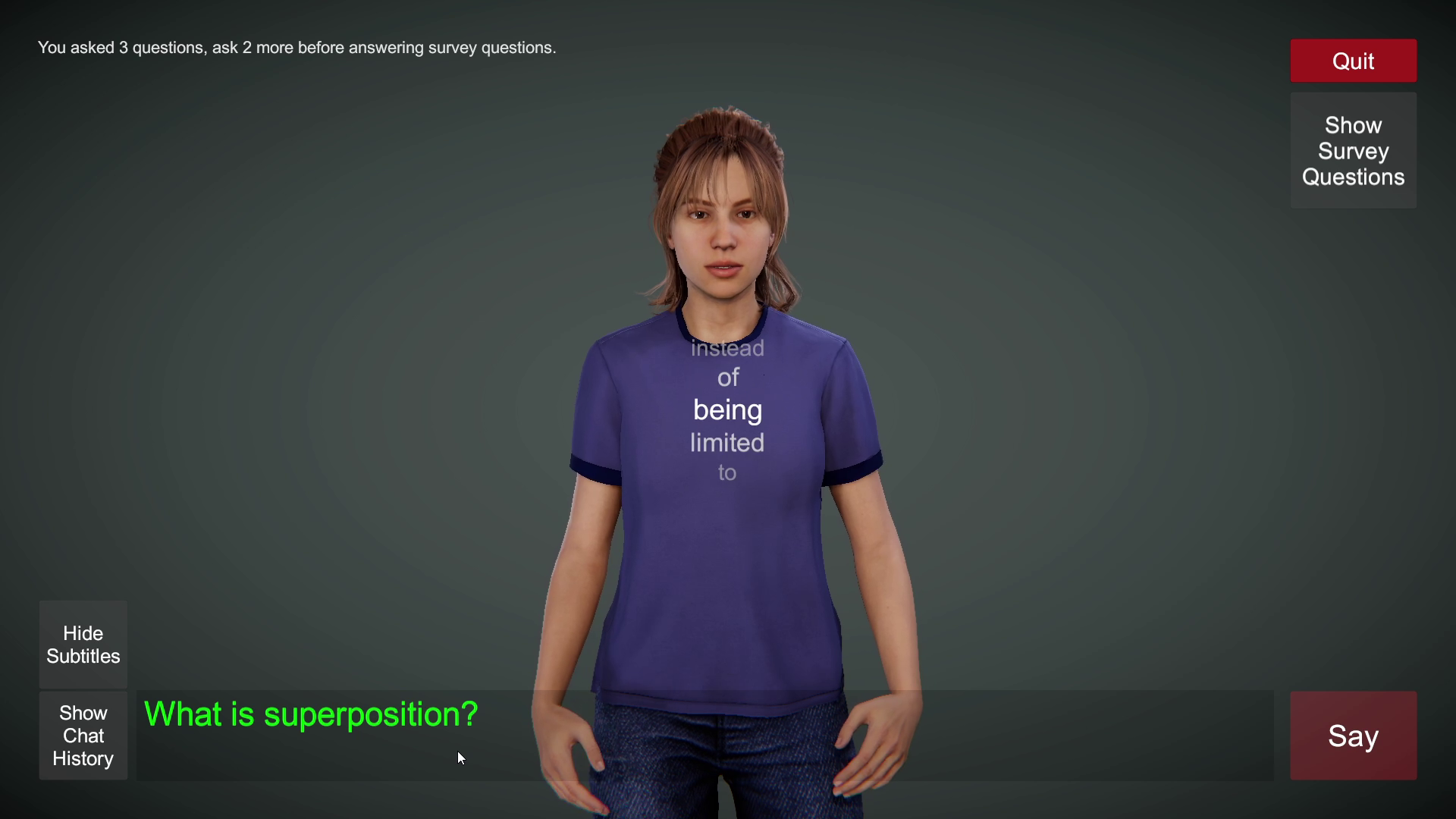}
     \caption{Model E-Low}
     \label{fig:model-e-low}
   \end{subfigure}
   %\hfill
   \begin{subfigure}{0.48\textwidth}
     \centering
     \includegraphics[width=\textwidth]{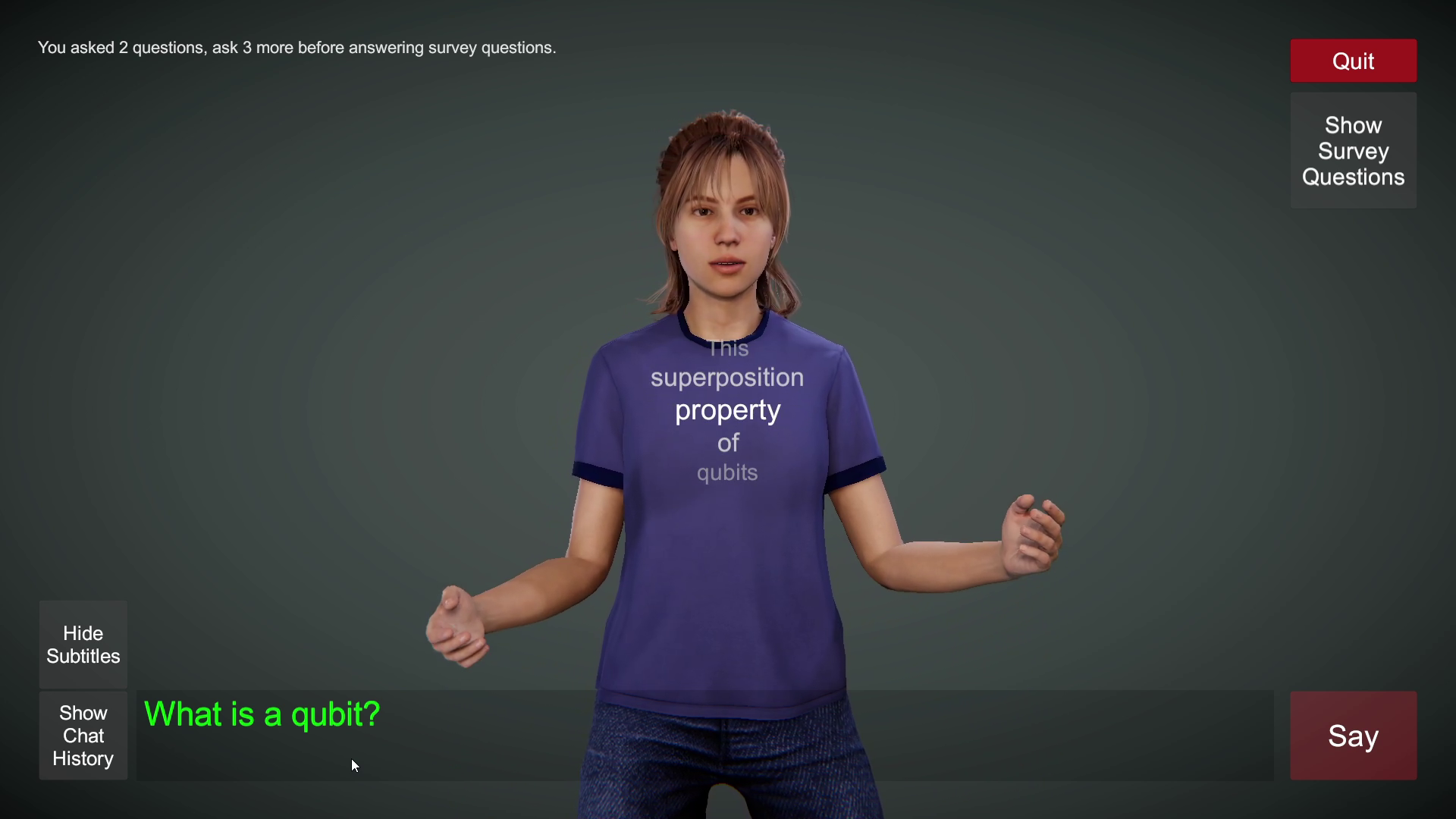}
     \caption{Model E-High}
     \label{fig:model-e-high}
   \end{subfigure}
    \caption{Sample screenshot from different models and their variations.}
    \label{fig:models}
\end{figure*}

Screenshots of different models are displayed in Figure~\ref{fig:models}. We name each variation of the system with its model name and whether they express high or low traits. For example, E-High refers to the variation where we express high trait personality using the model that uses both text and animation-based cues. Note that we display a single image for models A and D as they are visually similar in high and low variations. For the animation in Figure~\ref{fig:model-a}, the E-Low variation has the hands close to the body with a slightly more slanted posture (Figure~\ref{fig:model-e-low}), and the E-High variation has the hands further from the body with more upright posture (Figure~\ref{fig:model-e-high}).

\subsection{Study Design}
The study involved an educational application with a conversational agent that teaches users complex subjects. We presented participants with six options and asked them to select the least familiar topic. The topics were quantum computing, blockchain technologies, transformer architectures, quantum mechanics, string theory, and general relativity. The selected topic was provided to the GPT model as part of the system role prompt to guide a focused conversation. The application required that participants pose the agent at least five questions to learn about the topic. There was no upper limit on the number of questions a participant could ask. Upon completing their queries, participants could proceed to answer survey questions. They could review the survey questions or the chat history at any point.

The survey questions appeared in two groups. The first group included 27 questions on a 5-point Likert scale, where 15 questions measured the perceived personality of the agent using the extra-short form of the Big Five Inventory–2~(BFI-2-XS)~\cite{soto2017short}, and 12 questions measured self-assessment of learning, quality, and engagement using the Learning Object Evaluation Scale for Students~(\mbox{LOES-S})~\cite{kay2009assessing}. In LOES-S, learning-related questions are about the self-assessment of learning and how much the learning object, i.e., the tool in question, helped teach the subjects a new concept. Quality assesses the instructional design, ease of use, organization, and help features. Engagement evaluates how much the subjects liked the tool and whether they found it motivating.

The second group of questions required open-ended input to receive detailed participant feedback. There was no character limit for the answers to the open-ended questions. Completing both groups of questions directed the participants to the user study completion page, where they received a link for task approval.

\subsection{Participants}
We used the crowd-sourcing service Prolific to recruit participants. Before running the study, each participant was directed to a website to test whether they had installed the correct text-to-speech package. Only those with the supported system configurations could continue with the study. Two hundred ten unique participants (99 female, 95 male, 16 not specified) rated our system, with each alternative evaluated by 35 individuals, which provides a medium effect size (Cohen's $f = 0.26$) for both main effects and their interaction and power of $0.80$ at a significance level of $0.05$ for independent-subjects Analysis of Variance (ANOVA).

Each participant interacted with only one version of the system, where the average interaction time was $19.74\pm9.25$ minutes. This time excludes the introduction, during which participants read about the task and downloaded the application, but includes the time spent answering the survey questions. The average participant age was $28.80\pm8.57$. Upon entering their Prolific IDs, participants were shown an introduction message with information about the study details. We informed the participants about the data collected and the study's aim to measure the system's performance; we emphasized that the study did not aim to measure their knowledge in any manner.

\section{Quantitative Analysis}
\subsection{Data Organization and Exploratory Analysis}
BFI-2-XS includes three questions for each personality factor, half of which are inversely proportional to the measured dimensions. Responses were assigned integer values on a 5-point Likert scale, ranging from -2 to 2. We calculated the signed sum of these values to derive personality scores between -6 and 6, which were then re-scaled back to the range $[-2, 2]$. Similarly, LOES-S has five questions measuring learning, four questions measuring quality, and three questions measuring the engagement of the learning object. We calculated the sum per measurement type and mapped the corresponding ranges into $[-2, 2]$ to report the corresponding means.

\begin{figure*} 
\centering
\includegraphics[width=0.95\linewidth]{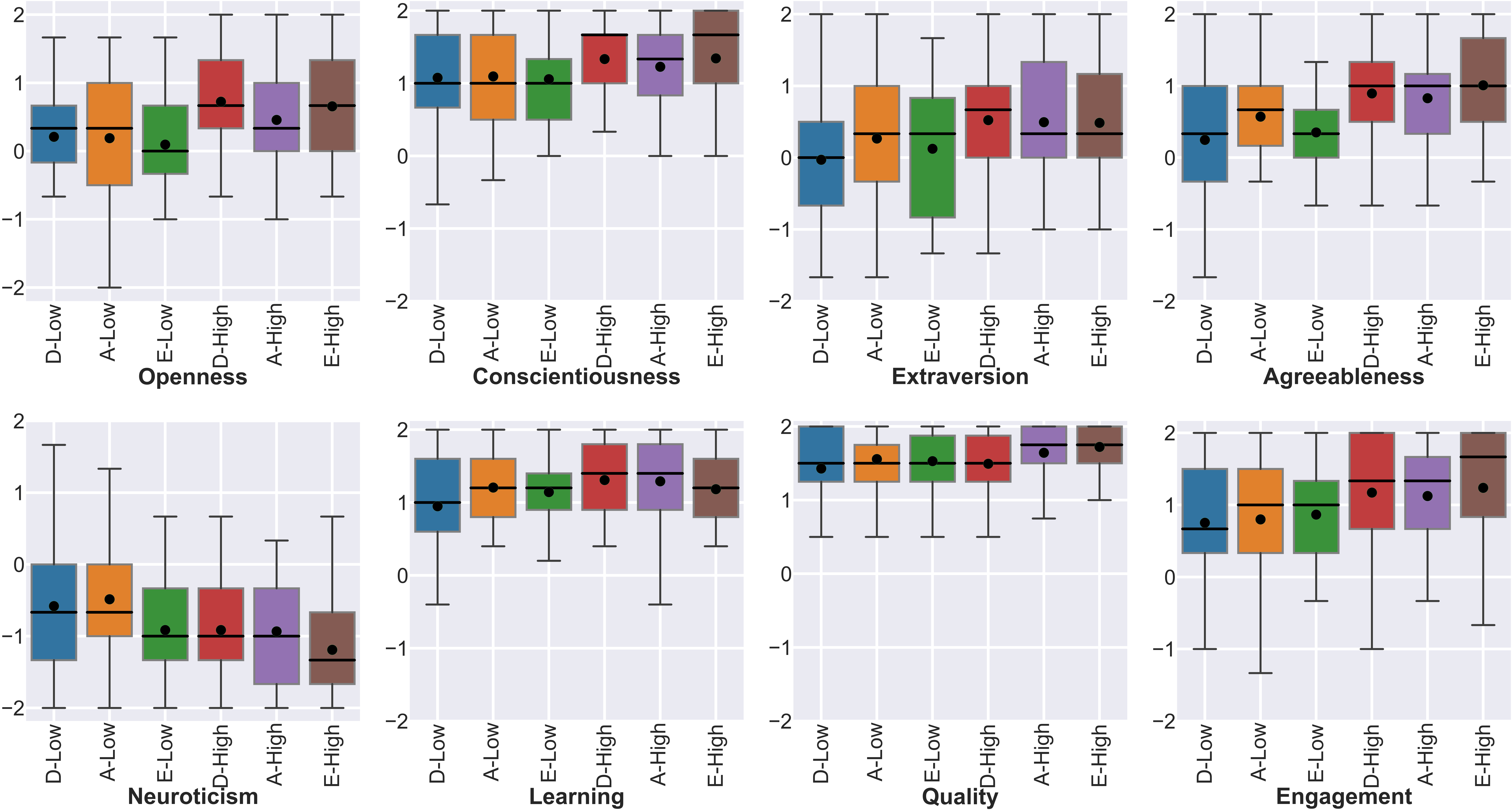}
\caption{Box plots of each variation's BFI-2-XS and LOES-S scores.}
\label{fig:box-plots}
\end{figure*}

For exploratory analysis, we display box plot diagrams of each model regarding perceived personality and LOES-S scores for learning, quality, and engagement (see~Figure~\ref{fig:box-plots}). The diagrams indicate positive mean scores for openness, conscientiousness, extroversion, and agreeableness and negative mean scores for neuroticism across all models. The models received particularly high ratings for conscientiousness.    The plots also show high positive ratings for learning, quality, and engagement, with mean engagement scores slightly higher for high personality variations than for low personality ones.  We can also observe that the model E-High, followed by A-High, represents high conscientiousness, high agreeableness, and low neuroticism better than the other models.  In the next section, we perform descriptive analysis to identify potential statistically significant effects of the models and personality styles on the output variables.

\subsection{The Effects of Model and Personality Styles on Learning Outcomes and Personality Perception}

To investigate the impact of model type (D, A, and E) and personality style (high or low) on learning outcomes and perceived personality factors, we ran seven two-way analysis of variance (ANOVA) models and one non-parametric alternative model (Welch’s ANOVA). Welch’s ANOVA was utilized to assess the quality scores of LOES-S across model type and personality style, given that the assumption of equal variances was violated, as indicated by a Bartlett test.

\begin{figure*}
 \centering
 \includegraphics[width=0.99\linewidth]{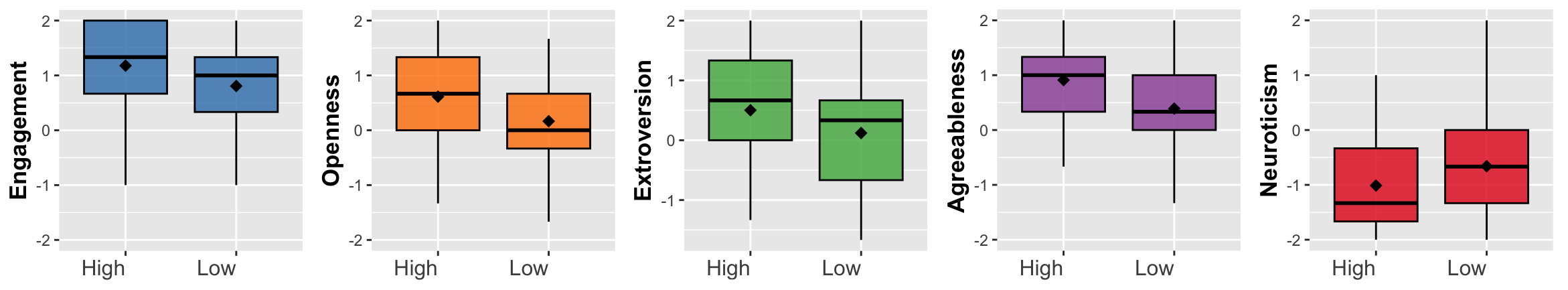}
 \caption{Box plots of statistically significant effects.}
 \label{fig:box-plots-significant}
\end{figure*}

Apart from the non-parametric model, which combined model type and personality style as a single factor, all other models examined the influence of model type and personality style on outcome mean individually and any potential interaction between these factors. With balanced and sufficiently large sample sizes ($n{=}35$) across factor combinations and evidence for equal variances across factor levels (as measured by Bartlett's test), all outcomes except for quality were appropriate for ANOVA modeling. To control the familywise error rate at $0.05$, we employed the Hommel method to adjust for multiple testing across all model terms, including one post-hoc analysis. Unlike the conservative Bonferroni correction, the Hommel method offers increased statistical power. Table~\ref{tab:sigOmnibus} presents significant terms for all ANOVA runs before and after the correction for multiple testing. Adjusted-for ANOVA tests returned significant main effects of personality style on the outcomes of engagement ($F = 9.502, p = 0.0421$), openness ($F = 4.474, p = 0.00178$), extroversion ($F = 10.148, p = 0.03006$), agreeableness ($F = 25.541, p = 0.0000211$), and neuroticism ($F = 9.708, p = 0.0378$). Although conscientiousness initially carried a significant finding for personality style ($F = 5.68, p = 0.0181$), this term's statistical significance dropped after multiple testing corrections. The main effect of the model type was initially significant for neuroticism, but the effect did not remain significant after the Hommel procedure ($p = 0.417$). The box plots of the statistically significant effects are depicted in Figure~\ref{fig:box-plots-significant}.

The effects of agent gender and skin color were not among the hypotheses, so we randomly selected one 3D agent model among four different appearances, which also determined the agent's voice, to support variety. We do not observe a significant effect due to the agent's gender or skin color, which confirms previous work~\cite{castro2021effectiveness}.

\begin{table}[ht]
 \setlength{\tabcolsep}{4.7pt}
 \def\arraystretch{1.5}
  \centering
  \caption{Two-way ANOVA significant findings for learning outcomes and perceived personality on model type and personality style ($n=210$). Statistically significant factors ($p<0.05$) after p-value adjustment are emphasized in bold.}
  \begin{tabular}{l|l|l|l|l}
     Outcome & Factor & F & p-value & Adj. p-value\\     
    \midrule
    \textbf{Engagement} & Pers. Style  & 9.502 & 0.002 & \textbf{0.042}\\
    \textbf{Openness} & Pers. Style  & 4.474 &  $<0.001$ & \textbf{0.002}\\
    Conscientiousness & Pers. Style &  5.68 & 0.018 & 0.290 \\
    \textbf{Extroversion} & Pers. Style  & 10.148 & 0.002 & \textbf{0.031}\\
    \textbf{Agreeableness} & Pers. Style & 25.541 & $<0.001$ & $<\textbf{0.001}$\\
    \textbf{Neuroticism} & Pers. Style  &  9.708 & 0.002& \textbf{0.038}\\
    Neuroticism & Model Type  &  3.681 & 0.027& 0.417\\   
  \end{tabular}  
  \label{tab:sigOmnibus}
  % \vspace*{-2.5ex}
\end{table}

\subsection{Correlations between Personality Factors and Learning Scores}
We report the Pearson Correlation between LOES-S and personality factors in Table~\ref{tab:pearson}. In general, correlation coefficients higher than $0.4$ are considered moderate. Perceived openness, conscientiousness, extroversion, and agreeableness all positively correlate to the LOES-S scores, albeit some weakly. The highest correlations are for conscientiousness, particularly for Model D. Quality and engagement scores are strongly correlated ($>0.6$), and learning is moderately correlated, nearing the threshold for a strong correlation. For Model D, openness and agreeableness also have moderate correlations with all the learning parameters. In general, engagement is moderately correlated with all factors except neuroticism. The expressed neuroticism is weakly inversely proportional to each parameter, but its only statistically significant correlations are for learning in Model D and quality in Model A. The correlations are the strongest for Model D and weakest for Model E.

\begin{table}[htbp]
\centering
\caption{Pearson correlation ($r$) of \textbf{Le}arning, \textbf{Qu}ality, and \textbf{En}gagement measurements of each model with each personality factor. $^{*}$ indicates $p < 0.05$, $^{**}$ indicates $p < 0.001$. The cell colors indicate transition from \colorbox[HTML]{440154}{\textcolor{white}{weak}} to \colorbox[HTML]{fde725}{strong} correlation.}
\begin{tabular}{c|c|ccccc}
\hline
\textbf{Model} & \textbf{Corr.} & \textbf{O} & \textbf{C} & \textbf{E} & \textbf{A} & \textbf{N} \\ \hline
\multirow{3}{*}{D} & \textbf{$r_{\textit{Le}}$} & \cellcolor[HTML]{35b779}\textcolor{black}{$.422^{**}$} & \cellcolor[HTML]{90d743}\textcolor{black}{$.582^{**}$} & \cellcolor[HTML]{21918c}\textcolor{white}{$.396^{**}$} & \cellcolor[HTML]{35b779}\textcolor{black}{$.446^{**}$} & \cellcolor[HTML]{21918c}\textcolor{white}{$-.349^{*}$} \\ \cline{2-7}
& \textbf{$r_{\textit{Qu}}$} & \cellcolor[HTML]{35b779}\textcolor{black}{$.437^{**}$} & \cellcolor[HTML]{fde725}\textcolor{black}{$.654^{**}$} & \cellcolor[HTML]{31688e}\textcolor{white}{$.279^{*}$} & \cellcolor[HTML]{21918c}\textcolor{white}{$.325^{*}$} & \cellcolor[HTML]{443983}\textcolor{white}{$-.190$} \\ \cline{2-7}
& \textbf{$r_{\textit{En}}$} & \cellcolor[HTML]{35b779}\textcolor{black}{$.417^{**}$} & \cellcolor[HTML]{fde725}\textcolor{black}{$.608^{**}$} & \cellcolor[HTML]{31688e}\textcolor{white}{$.287^{*}$} & \cellcolor[HTML]{35b779}\textcolor{black}{$.422^{**}$} & \cellcolor[HTML]{443983}\textcolor{white}{$-.109$} \\ \hline
\multirow{3}{*}{A} & \textbf{$r_{\textit{Le}}$} & \cellcolor[HTML]{21918c}\textcolor{white}{$.351^{*}$} & \cellcolor[HTML]{21918c}\textcolor{white}{$.348^{*}$} & \cellcolor[HTML]{35b779}\textcolor{black}{$.456^{**}$} & \cellcolor[HTML]{21918c}\textcolor{white}{$.398^{**}$} & \cellcolor[HTML]{31688e}\textcolor{white}{$-.208$} \\ \cline{2-7}
& \textbf{$r_{\textit{Qu}}$} & \cellcolor[HTML]{31688e}\textcolor{white}{$.215$} & \cellcolor[HTML]{21918c}\textcolor{white}{$.378^{*}$} & \cellcolor[HTML]{31688e}\textcolor{white}{$.204$} & \cellcolor[HTML]{443983}\textcolor{white}{$.150$} & \cellcolor[HTML]{35b779}\textcolor{black}{$-.425^{**}$} \\ \cline{2-7}
& \textbf{$r_{\textit{En}}$} & \cellcolor[HTML]{90d743}\textcolor{black}{$.504^{**}$} & \cellcolor[HTML]{21918c}\textcolor{white}{$.332^{*}$} & \cellcolor[HTML]{90d743}\textcolor{black}{$.505^{**}$} & \cellcolor[HTML]{90d743}\textcolor{black}{$.541^{**}$} & \cellcolor[HTML]{31688e}\textcolor{white}{$-.233$} \\ \hline
\multirow{3}{*}{E} & \textbf{$r_{\textit{Le}}$} & \cellcolor[HTML]{31688e}\textcolor{white}{$.266^{*}$} & \cellcolor[HTML]{35b779}\textcolor{black}{$.405^{**}$} & \cellcolor[HTML]{31688e}\textcolor{white}{$.281^{*}$} & \cellcolor[HTML]{443983}\textcolor{white}{$.184$} & \cellcolor[HTML]{440154}\textcolor{white}{$-.046$} \\ \cline{2-7}
& \textbf{$r_{\textit{Qu}}$} & \cellcolor[HTML]{21918c}\textcolor{white}{$.334^{*}$} & \cellcolor[HTML]{443983}\textcolor{white}{$.133$} & \cellcolor[HTML]{440154}\textcolor{white}{$.056$} & \cellcolor[HTML]{21918c}\textcolor{white}{$.305^{*}$} & \cellcolor[HTML]{440154}\textcolor{white}{$-.054$} \\ \cline{2-7}
& \textbf{$r_{\textit{En}}$} & \cellcolor[HTML]{35b779}\textcolor{white}{$.401^{**}$} & \cellcolor[HTML]{21918c}\textcolor{white}{$.353^{*}$} & \cellcolor[HTML]{21918c}\textcolor{white}{$.389^{**}$} & \cellcolor[HTML]{21918c}\textcolor{white}{$.353^{*}$} & \cellcolor[HTML]{440154}\textcolor{white}{$-.052$} \\ \hline
\end{tabular}
\label{tab:pearson}
\end{table}

\section{Qualitative Analysis}
% This section presents the qualitative analysis of the answers to the open-ended questions after the conversation with agents.

\subsection{Data}
We posed the following open-ended questions to the participants for in-depth feedback:

\begin{enumerate}
    \item What is \textit{$<$topic$>$} based on your conversation with the system?
    \item Why is \textit{$<$topic$>$}important, please discuss briefly?
    \item Did you learn anything new about \textit{$<$topic$>$} after your conversation with the system? If so, what did you learn?
    \item Do you think the system's behavior influenced your learning experience?
    \item Please briefly describe your conversation with the system. Was it interesting? Did you encounter any problems?
\end{enumerate}

Two independent researchers tagged the answers with at least 95\% agreement for each theme. The themes were determined based on the following criteria:

\begin{itemize}
    \item[\textbf{1}] \textbf{Benefit/No Benefit to Learning Experience:} Regarding the third and fourth questions, this theme reflects the effect of the system behavior on the participant's learning experience.
    \item[\textbf{2}] \textbf{Learning / No Learning:} Regarding the third question, this theme captures whether the participant learned anything new interacting with the system.   
    \item[\textbf{3}] \textbf{Interesting / Not Interesting:} Regarding the fifth question, this theme considers whether the participant finds the system interesting or not.
    \item[\textbf{4}] \textbf{Realistic / Robotic / Poor Voice:} This theme captures the overall human likeness of the system based on the answers to the fourth and fifth questions. In certain cases, the participants only commented about the system's voice, which utilized less developed local solutions to support immediate response; otherwise, the participants' feedback is on the overall image of the agent.
    \item[\textbf{5}] \textbf{Detailed / Moderate / Short Answer:} Regarding the answers to the first and second questions, this theme captures how much detail the response includes. Although this may strongly depend on the participants' characteristics, the system's behavior may have also caused them to adjust the level of detail of their answers. These answers may also depend on the participant's learning; the less they know, the shorter the answers will be. We tag the responses as detailed or short; any remaining response is considered moderate.
\end{itemize}

Table~\ref{tab:theme} shows the number of themes for each variation. The following subsections focus on each theme with quotes from the participants. At the end of each quote, we indicate the participant's ID as a reference and the variation that the participant interacted with. We only fixed the typos in some of the responses to improve readability.

\begin{table}[ht]
\setlength{\tabcolsep}{2.2pt}
\centering
\footnotesize
\caption{Theme analysis results where the numbers depict the occurrence of each theme for each variation. The cell colors indicate transition from \colorbox[HTML]{440154}{\textcolor{white}{low}} to \colorbox[HTML]{fde725}{high} over 35 participant answers per variation.}
\begin{tabular}{|l|c|c|c|c|c|c|}
\hline
\textbf{Variation} & \textbf{D-Low} & \textbf{D-High} & \textbf{A-Low} & \textbf{A-High} & \textbf{E-Low} & \textbf{E-High} \\ \hline
Benefit & \cellcolor[HTML]{21918c}\textcolor{white}{$15$} & \cellcolor[HTML]{35b779}\textcolor{white}{$19$} & \cellcolor[HTML]{35b779}\textcolor{white}{$17$} & \cellcolor[HTML]{90d743}\textcolor{black}{$22$} & \cellcolor[HTML]{35b779}\textcolor{white}{$19$} & \cellcolor[HTML]{90d743}\textcolor{black}{$21$} \\ \hline
No Benefit & \cellcolor[HTML]{31688e}\textcolor{white}{$9$} & \cellcolor[HTML]{443983}\textcolor{white}{$4$} & \cellcolor[HTML]{31688e}\textcolor{white}{$8$} & \cellcolor[HTML]{443983}\textcolor{white}{$6$} & \cellcolor[HTML]{31688e}\textcolor{white}{$8$} & \cellcolor[HTML]{443983}\textcolor{white}{$7$} \\ \hline
Learning & \cellcolor[HTML]{90d743}\textcolor{black}{$23$} & \cellcolor[HTML]{fde725}\textcolor{black}{$25$} & \cellcolor[HTML]{90d743}\textcolor{black}{$21$} & \cellcolor[HTML]{fde725}\textcolor{black}{$26$} & \cellcolor[HTML]{fde725}\textcolor{black}{$25$} & \cellcolor[HTML]{fde725}\textcolor{black}{$29$} \\ \hline
No Learning & \cellcolor[HTML]{443983}\textcolor{white}{$6$} & \cellcolor[HTML]{443983}\textcolor{white}{$6$} & \cellcolor[HTML]{443983}\textcolor{white}{$5$} & \cellcolor[HTML]{440154}\textcolor{white}{$2$} & \cellcolor[HTML]{443983}\textcolor{white}{$7$} & \cellcolor[HTML]{440154}\textcolor{white}{$2$} \\ \hline
Interesting & \cellcolor[HTML]{21918c}\textcolor{white}{$15$} & \cellcolor[HTML]{90d743}\textcolor{black}{$22$} & \cellcolor[HTML]{35b779}\textcolor{white}{$19$} & \cellcolor[HTML]{90d743}\textcolor{black}{$20$} & \cellcolor[HTML]{90d743}\textcolor{black}{$21$} & \cellcolor[HTML]{90d743}\textcolor{black}{$20$} \\ \hline
Not Interesting & \cellcolor[HTML]{440154}\textcolor{white}{$3$} & \cellcolor[HTML]{440154}\textcolor{white}{$2$} & \cellcolor[HTML]{440154}\textcolor{white}{$0$} & \cellcolor[HTML]{440154}\textcolor{white}{$1$} & \cellcolor[HTML]{440154}\textcolor{white}{$2$} & \cellcolor[HTML]{440154}\textcolor{white}{$1$} \\ \hline
Realistic & \cellcolor[HTML]{440154}\textcolor{white}{$0$} & \cellcolor[HTML]{440154}\textcolor{white}{$3$} & \cellcolor[HTML]{440154}\textcolor{white}{$2$} & \cellcolor[HTML]{440154}\textcolor{white}{$3$} & \cellcolor[HTML]{440154}\textcolor{white}{$1$} & \cellcolor[HTML]{443983}\textcolor{white}{$4$} \\ \hline
Robotic & \cellcolor[HTML]{440154}\textcolor{white}{$1$} & \cellcolor[HTML]{440154}\textcolor{white}{$1$} & \cellcolor[HTML]{443983}\textcolor{white}{$6$} & \cellcolor[HTML]{443983}\textcolor{white}{$5$} & \cellcolor[HTML]{443983}\textcolor{white}{$5$} & \cellcolor[HTML]{440154}\textcolor{white}{$3$} \\ \hline
Poor Voice & \cellcolor[HTML]{31688e}\textcolor{white}{$9$} & \cellcolor[HTML]{443983}\textcolor{white}{$5$} & \cellcolor[HTML]{440154}\textcolor{white}{$3$} & \cellcolor[HTML]{443983}\textcolor{white}{$6$} & \cellcolor[HTML]{440154}\textcolor{white}{$2$} & \cellcolor[HTML]{440154}\textcolor{white}{$3$} \\ \hline
Detailed Answer & \cellcolor[HTML]{440154}\textcolor{white}{$2$} & \cellcolor[HTML]{31688e}\textcolor{white}{$9$} & \cellcolor[HTML]{440154}\textcolor{white}{$2$} & \cellcolor[HTML]{443983}\textcolor{white}{$6$} & \cellcolor[HTML]{31688e}\textcolor{white}{$8$} & \cellcolor[HTML]{443983}\textcolor{white}{$4$} \\ \hline
Moderate Answer & \cellcolor[HTML]{31688e}\textcolor{white}{$10$} & \cellcolor[HTML]{31688e}\textcolor{white}{$10$} & \cellcolor[HTML]{31688e}\textcolor{white}{$10$} & \cellcolor[HTML]{31688e}\textcolor{white}{$8$} & \cellcolor[HTML]{31688e}\textcolor{white}{$10$} & \cellcolor[HTML]{90d743}\textcolor{black}{$20$} \\ \hline
Short Answer & \cellcolor[HTML]{90d743}\textcolor{black}{$23$} & \cellcolor[HTML]{35b779}\textcolor{white}{$16$} & \cellcolor[HTML]{90d743}\textcolor{black}{$23$} & \cellcolor[HTML]{90d743}\textcolor{black}{$21$} & \cellcolor[HTML]{35b779}\textcolor{white}{$17$} & \cellcolor[HTML]{31688e}\textcolor{white}{$11$} \\ \hline
\end{tabular}
\label{tab:theme}
\end{table}

\subsection{Benefit}
A primary objective of agent-based educational systems is to facilitate self-directed learning by providing a useful learning experience.  Actively involving individuals in learning helps them develop a deeper understanding of the subject matter than passive learning through only reading or watching the material~\cite{cairncross2001interactive}. This level of interaction is naturally present with a human teacher who can promptly address a learner's questions. We required participants to actively engage by asking questions to replicate this dynamic and prevent passive learning.

Most participants indicated that the system behavior improved their learning experience; they emphasized increased engagement due to having a human-like agent: \textit{``It was more engaging to have a human avatar instead of a blank screen or other representation.'' -- P3 (A-High).} Even for the dialogue-only variations, the experience was more enjoyable due to interaction: \textit{``I think I enjoyed learning using the system more than I would have if I were reading on my own in a book or Google.'' -- P21 (D-High).} Being able to ask questions using natural language and receive to-the-point answers was also found to be beneficial by certain participants: \textit{``It saved me some time from having to Google specific terms and read long texts on them, giving me the key points to get a basic understanding.'' -- P17 (E-High).} A human-like interaction with a non-human system can help people with anxiety to experience interactive learning: \textit{``It can be useful a lot, and I'd like to use it because it is quite calming down the person who has anxiety.'' -- P27 (E-High).} 

However, a human teacher would typically simplify the material to facilitate easier learning, which can be challenging for a language model to emulate with limited user feedback:  \textit{``I believe the system was very informational, which influenced my learning. However, the information was too much to handle at once, which clouded my mind.'' -- P30 (A-High).} Participants who reported no benefit from the system often cited the difficulty of the answers: \textit{``I could have learned more on Wikipedia or Google. The system's answers were too difficult to understand for me.'' -- P169 (D-Low).}   Although the high-trait models were generally found to be more beneficial, some participants noted that their responses were too lengthy: \textit{``The responses were quite long, sometimes too long. Some of the things it said could've been left out as they didn't provide any useful information; it was just 'flavor text.' I also read very fast, so waiting for it to stop talking was a bit boring. Other than that, it was a very positive experience. '' -- P42 (D-High).}

In addition to length, a major difference between the high and low variations was the LLM's word choices. The high-trait variations use motivational language, supporting the learner instead of just giving the answer: \textit{``Yes, the system had a motivational tone that lifted my energy towards learning about something I had zero knowledge about.'' -- P37 (D-High).} Such language can help create a more interactive and thus motivational experience: \textit{``Yes, I liked the easy-to-learn explanations and also the motivational part 'that is indeed a fantastic question'.'' -- P106 (D-High).} Participants also referred to the system as friendly (P50 and P150, A-High) and polite (P209, D-High). On the other hand, the low-trait variations responded with shorter sentences that some users preferred: \textit{``It was not interesting per se; however, it was very informative and straight to the point.'' -- P2 (E-Low).} Especially, the introverted language~\cite{mairesse2009can} of the low variations could influence the relatively short and simple responses: \textit{``Conversation was simple and quick. The agent gave me short and simple answers to my question that are easy to understand'' -- P170 (E-Low).}

Participants found variations with embodied agents slightly more beneficial and interesting. Especially, they  reported a positive influence of having an embodied human agent for the high-trait variations:  \textit{``Somewhat, seeing a human-like face made it easier to memorize the information'' -- P163 (A-Low).} \textit{``Interacting with a humanoid entity is more engaging than reading a book.'' -- P143 (A-High).}  \textit{``Yes! I really enjoyed it; it seemed very human. It was like talking to an expert; it can answer any question you have instantly.'' -- P12 (A-High).}  \textit{``It was really exciting to see a person/character in front of the screen. Of course, it has an influence and really affected my learning process positively.'' -- P134 (E-High).}   \textit{``I think the animations and the looks of the character were motivating, and this could help with the learning in general.'' -- P207 (E-High).} \textit{``Body movement and text to speech allowed to be more engaged in the conversation.'' -- P184 (E-High).} A human-like agent can help learners better focus on the conversation, positively affecting learning: \textit{``Having someone explaining a subject to you in human form generates a curiosity that is similar to listening to an enthusiastic teacher. As someone with a low attention span, the agent kept me engaged in the conversation and sparked further interest.'' -- P1 (E-High).}   \textit{``It was really exciting to see a person/character in front of the screen. Of course, it has an influence and really affected my learning process positively.'' -- P134 (E-High).} While most participants focused on the body movements, a few reflected on the facial expressions and their positive effect in the E-High variant: \textit{``Slight facial ``expressions'' was noted and kind of felt like it made an impact, to be fair.'' -- P11 (E-High).}

For the low-trait agents, participants noted that agent movements were monotonic and the visual representation brought no advantage: \textit{``The person itself is extremely dull, there is no life if that makes sense, and the movements and gestures are extremely weird, the hand and arm movements are strange, I would rather have that taken away, but being able to ask any questions to a topic and a response provided immediately is amazing, really like that aspect.'' -- P196 (E-Low).}  \textit{``Maybe, I think if the system were "nicer" and less monotonic, the learning would be easier.'' -- P147 (E-Low).}  \textit{``I did not find the ``graphics'' to help. A chatbot would have basically had the same effect on me.'' -- P191 (A-Low).}

\subsection{Learning}

The results suggest that high-trait personality variations lead to improved learning outcomes. However, the influence of expressivity appears limited, as Model E shows only a slight improvement in learning performance compared to Model A. Only two responses in each A-High and E-High were associated with no learning, pointing out that human-like appearance combined with high traits can improve learning. 

The system sometimes inspired participants to learn more about the topic: \textit{``I did not know anything about this theory at all. After my conversation, I can proudly say that I am really into string theory. I learned the basic concept and the creators of the theory. I also asked how I could learn more, and the conversational agent suggested four different possible sources.'' -- P8 (A-High).} \textit{``It really almost felt like talking to someone who knows quantum computing well.  I especially appreciated the way it understood my questions, even though I felt a question or two were a bit vague.  The system actually made me want to know more about the subject so I can ask better questions.'' -- P11 (E-High).} Since we asked participants to choose the subject they had the least information about, most of them reported having almost no prior knowledge of the conversation topic: \textit{``It was a completely new topic to me, and I feel generally satisfied with the answers of your system'' -- P15 (E-Low).} Therefore, the themes associated with learning correspond to the participants who are newly introduced to the subject and acquired new information.  A few participants indicated they already had sufficient knowledge of the subject: \textit{``I didn't learn anything new; I already knew the basic information about Blockchain.'' -- P29 (E-Low).} Still, such participants found the system useful and reported a desire to use it again: \textit{``I am an enthusiast of general relativity. I honestly didn't learn much, but I would love to use the AI to learn more in the future!'' -- P101 (E-Low).}

Participants categorized under the ``no learning'' theme generally indicated the difficulty of the subject: \textit{``Previously, I had no idea what transformer architecture was. But unfortunately, I still believe that I did not learn a great deal about this type of technology. In my opinion, these new technologies that use AI are very difficult to understand if a person has no background knowledge.'' -- P20 (E-High).} One detriment to learning could be the agent's short answers in the low-trait variations: \textit{``Not much (learning) as the replies were brief, but I got a basic idea'' -- P101 (E-Low).} Conversely, the lengthy answers of the high-trait variations may have been distracting: \textit{``The topics answered were on point, maybe a bit too long, and different questions had similar answers in common as part of it.'' -- P41 (E-High).}

\subsection{Interestingness and Realism}
Promoting interest in a subject is essential for success in education~\cite{harackiewicz2016interest}. For conversational agents, behavioral and graphical realism is an important factor for engagement~\cite{latoschik2017effect}, which also applies to pedagogical agents~\cite{salehi2019effect}.

Participants found the D-Low variation to be the least interesting, followed by A-Low. All the high-trait variations and E-low were perceived as similarly interesting. The expressive gesturing in the E-Low variation may have mitigated the decrease in interest. A few participants found the agent realistic in all variations, with the E-High variation receiving the most responses associated with the ``realistic'' theme.

Participants who found the system interesting usually reported a positive influence on learning in the E-High variation: \textit{``I found it fascinating, really interesting, and quickly increased my knowledge on the subject. I would have loved to do more and carry on asking questions to discover more about blockchain.'' -- P199 (E-High).} \textit{``It was really exciting to see a person/character in front of the screen. Of course, it has an influence and really affected my learning process positively.'' -- P134 (E-High).} \textit{``I think that the system behaved as naturally as possible, so it was a good influence.'' -- P174 (E-High).} Participants mentioned a positive effect on concentration due to the existence of the human avatar: \textit{``...Avatar of the agent helps concentrate on the conversation.''} -- P170 ( E-Low).  The experience's novelty could have resulted in some participants finding the study interesting: \textit{``It was quite interesting. I did not know what to expect when entering the task, but I was pleasantly surprised and engaged in the entire experience. It would definitely be something I would use again if I could.'' -- P175 (A-High).} Participants found Model D less interesting but still beneficial in teaching: \textit{``I think it wasn't interesting, but it taught me a topic I didn't know about.'' -- P2 (D-Low).} Since our focus was on a short conversation with the agent, the users could have endured the negative aspects of Models~D and A more easily. The positive outcomes of expressive gestures may become more prominent in the long term. For example, keeping the agent's behavior more interesting can be important to keep the user's attention longer.

The system's voice was highly criticized due to using a local solution for immediate responses at the cost of realism: \textit{``I didn't like the voice because it seemed robotic and emotionless.'' -- P10 (D-Low).}  Some participants found the speech inhuman and slow: \textit{``I prefer the more traditional 'read what the bot says' instead of it talking with its inhuman voice that speaks way slower and more monotonously than I would prefer.'' -- P41 (E-High).} There was an increased focus on the negative aspects of speech synthesis when there was no visual representation, with 14 participants mentioning the poor voice of the agent in Model D. In contrast, negative comments on the agent's voice were as low as 5 in Model E.  Since Model D does not include an embodied agent, only a few participants regarded the agent as a robotic individual, with most of the negative comments focusing on the voice.  Among the variations with a visual representation, E-High had the fewest responses associated with the ``robotic'' theme, which can be interpreted as the high-trait expressive motion effectively enhancing the agent's realism.

Participants who found Model D human-like mostly reflected on the naturalness of the generated responses: \textit{``It was interesting. I liked the way it started to answer. It always added a kind of human comment and then gave you the answer.'' -- P59 (D-High).} Participants assigned to the embodied representations tended to describe the agent as a real human:\textit{``It felt like a conversation with a real human, not like a robot. I felt like he was my teacher; I could ask any question.'' -- P6 (A-Low).} \textit{``I really enjoyed it, it seemed very human.'' -- P12 (A-High).} \textit{``I think it gave a more human experience and helped me focus on the subject.'' -- P179 (E-Low).} The lack of expressive facial expressions in Model A was noticed by participants, which could explain why some found it robotic: \textit{``Some facial expressions might make the bot more relatable.'' -- P163 (A-Low).} One participant found the overall appearance of the agent uncanny: \textit{``At first, I was surprised that this 'agent' has a human-like avatar. I don't think it helped me through the learning process because it felt kinda uncanny. Also, the TTS quality was poor, so it would be hard to understand some words if not for the subtitles. But I think that agent's responses were very good and comprehensive.'' -- P57 (E-High).}

\subsection{Answer Depth}
The depth of participants' answers to the first and second questions can indirectly measure their learning and attention. Therefore,  we grouped the responses into three categories based on how much detail they captured.  Short answers usually only summarize the conversation topic with a single sentence: \textit{``1. Quantum physics is a fundamental branch of physics that deals with the behavior of very small particles. 2. It provides information about the little particles, which are everywhere.'' -- P56 (E-Low).} Moderate-length answers use examples from the conversation and include more detail on the subject: \textit{``1. Blockchain is a decentralized technology that records transactions. It consists of blocks of data linked together in a chain. 2. It is important because of its transparency that allows one to view the transaction history, it is also highly secure.'' -- P103 (A-High).} Detailed answers contain more than one sentence for each question and include different aspects of the subject: \textit{``1. Blockchain is a decentralized system to store information about transactions. A single block contains data about the transaction and the hash of the previous block, so it can't be removed without breaking the chain. Data are stored as copies on many computers called nodes. 2. Blockchain is important because it is decentralized, and there is no authority keeping watch on it, so it is hard to manipulate it. It is also very secure, as every transaction must be validated by miners before it becomes a part of the chain.'' -- P57 (E-High).}

More participants provided detailed answers in the high-trait variations for Models D and A, possibly because the agent's longer responses encouraged them to elaborate. However, more participants gave detailed responses in the low-trait variation for Model E compared to its high-trait variation.  Notably, the E-High variation had the highest number of moderate-length responses, twice as many as the other models. Although the amount of detail in participant responses may depend on various factors, including their personality and attitude toward the experiment, we note a general trend toward lengthier responses for high-trait agents.

\section{Discussion}

All the agents received positive mean personality ratings across all traits except for neuroticism. Since neuroticism is the inverse of emotional stability, we can conclude that, regardless of modality and personality expression, the agents were perceived positively as open, conscientious, extroverted, agreeable, and emotionally stable.  Participant responses to open-ended questions also support this finding. The whole experience was perceived favorably even when low-trait personality variants, which were supposed to be less friendly, were employed.  Previous studies show that people find interactions with virtual agents engaging, informative, and usable~\cite{gratch2007can}.  However, the positive responses could also be due to the ``novelty effect'', an initial fascination with new technology. To mitigate such effects, techniques like extended tutorials and adaptive strategies can be employed~\cite{miguel2024evaluation}.

Although mean ratings were positive for both, high-trait agents received higher scores than low-trait agents for all the positive personality factors.   High-trait personality styles were associated with increased openness, extroversion, agreeableness, and emotional stability ratings with statistically significant effects.  The only factor that did not have a statistically significant relationship with style variation after multiple hypothesis testing was conscientiousness. Thus, for R1, we can conclude that personality style affects the perception of all the personality factors except conscientiousness.  Among these, agreeableness had the highest effect size, followed by extroversion.  This finding also helps validate the personality style expression adjustments in the system and the mappings of the LMA factors of Space, Weight, and Flow to extroversion and agreeableness. Participants' answers to open-ended questions suggest that they cared about the virtual agent's  `friendliness'' and ``niceness'' or lack thereof. These results align with the previous reports that students generally prefer teachers high in extroversion, agreeableness, and conscientiousness~\cite{tan2018students}. The variance in conscientiousness is difficult to discern in a short scenario~\cite{sonlu2021conversational}; so, the lack of statistically significant effects of style on its perception is expected. However, it is important to note that conscientiousness received the highest scores as the perceived agent personality, which may imply a tendency to attribute reliability and organizational skills to educational agents.

Regarding RQ2, we found no statistically significant effect of model type on perceived personality.  Similarly, for RQ3, no statistically significant effects of model type on LOES-S scores were observed. Thus, H1 was rejected as the absence or presence of visual representations did not impact learning outcomes assessed via LOES-S scores. Some participants indicated an indifference toward the graphical representation in their comments.  However, overall thematic analysis suggests that models A and E provided greater benefits on the learning experience, were found more interesting, and enhanced learning of a new topic compared to model D. This is in line with the literature indicating that the visual representation of an agent can enhance motivation, interest in the topic, and belief in its utility~\cite{baylor2009promoting}.  

For comparisons between high and low-trait styles, the evidence partially supported H2. Although both personality styles were positively rated across all models, the only significant difference was in the mean engagement scores.  High-trait agents were found more engaging than low-trait agents, supporting H2c. Participant responses to open-ended questions also confirm this finding. The absence of statistically significant differences in quality and learning scores of LOES-S across high and low-trait variants can be attributed to individual learning preferences. In their responses, some participants praised the directness of the low-trait agents, while others emphasized that the high-trait versions were motivational.  Such preferences may be linked to participant personalities and learning style preferences~\cite{furnham2005individual}.

For RQ4, we found positive correlations between perceived positive personality factors (O, C, E, A) and most learning outcomes for all the models. We also found a negative correlation between perceived neuroticism and learning for Model D and quality for Model A. The highest correlations between learning outcomes and perceived personality traits were for Model D, followed by Models A and E. Among all personality dimensions, conscientiousness yielded the highest correlation with learning outcomes. This is particularly evident in Model D, which suggests that the lack of a visual representation may have allowed participants to focus more on the educational aspects of the system.  Organizational skills describe Conscientiousness and is strongly associated with educational proficiency~\cite{kim2016conscientiousness}. Without a humanoid body that competes for attention, participants may have perceived the system as more focused and reliable in delivering educational content.

\section{Limitations and Future Work}
The LOES-S measures learning outcomes through self-reports, which do not necessarily reflect whether participants have effectively comprehended the material. Since our study did not assess the participants' mastery of the subject, we cannot confirm whether model variations and personality expression influenced learning performance. Although we asked open-ended questions about the subject, participants' responses were generally short and too vague to gauge comprehension accurately.  Future research could include pre- and post-assessment questions to measure learning more effectively.  Also,  because the system does not track comprehension, the complexity of the returned responses is not customized; therefore, some participants found them to be too long and difficult to understand.  Future work can utilize natural feedback, such as gaze~\cite{arslanyilmaz2023eye}, to track the learner's attention and understanding, which can help adjust the subject's difficulty. Similarly, nonverbal behaviors such as head movement and facial expressions can be used to estimate the learner's interest~\cite{nakamura2012investigation}. We believe that more accurate information can be achieved through long-term studies and multiple sessions during which learners have more opportunities to interact and familiarize themselves with the agent. 
% Learning by experience has long-term benefits~\cite{drissner2014short}, interacting with a human-like agent can have similar outcomes 

A major source of criticism about the system was the unnaturalness of the synthesized voice, as we used the operating system's built-in text-to-speech program for efficiency. The local solution had another disadvantage: the participants had to download the system to their computers and run it locally. In the future, we would like to provide text-to-speech as a built-in component and serve the tool as a web-based platform to reach a broader audience.  In this study, we kept the voice selections limited to one male and one female voice.  Experimenting with different voices and analyzing their effect on user perception could be a research direction. 

The thematic analysis indicates that individual differences across user preferences were prominent. A future direction is to employ agents customized for each user so that they have compatible personalities. Such selections can potentially increase the user’s trust and willingness to listen~\cite{zhou2019trusting}. In addition, we only tested two opposite personality combinations to keep the study duration and number of conditions feasible. More nuanced variations and incorporating other personality factors, such as conscientiousness, could provide more detailed information. 

A possible extension to our system is to adaptively generate agent animations to fit the generated dialogue. Currently, we use the same set of gestures to complement speech. Online gesture synthesis techniques that take text or audio as input can help overcome the problems related to naturalness, thus increasing engagement and learning quality.

\section{Conclusion}

This paper presents a conversational system and user study to explore how personality perception and embodiment affect learning outcomes.  Using GPT 3.5 and realistic 3D human models,  we created agents expressing high and low agreeableness and extroversion variations through dialogue and animation cues.  We designed three types of agents: a disembodied agent expressing personality through dialogue, one expressing personality only through dialogue, and an embodied agent expressing personality through dialogue and animation. We conducted a three-by-two independent-subjects user study with three agent models and the two personality variations, where each participant was asked to converse with an agent on a complex subject to learn about it. After the conversation, participants rated their version of the system based on their perceived personality of the agent and learning, quality, and engagement of the learning experience. The results indicate that regardless of the model choice, the whole experience was rated favorably in general, and participants judged the agents as high in openness, conscientiousness, extroversion, agreeableness, and low in neuroticism.  However, the degree of positive perception was lower in low-trait personality styles than in high-trait ones. Although the engagement score was higher for the embodied agent with expressive animations, we found no significant differences across the models for other learning outcomes. We hope that the findings of this work inspire future studies to utilize expressive animation and dialogue cues to improve the overall experience of educational applications with conversational agents. 

\balance

\section*{Acknowledgments}
This research is supported by The Scientific and Technological Research Council of Turkey (T\"{U}B\.{I}TAK) under Grant No.~122E123.

\end{document}